\documentclass[letterpaper, 10 pt, conference]{ieeeconf}
\IEEEoverridecommandlockouts
\usepackage{cite}
\usepackage{amsmath,amssymb,amsfonts}
\usepackage{xfrac}
\usepackage{algorithmic}
\usepackage{graphicx}
\usepackage{textcomp}
\usepackage{xcolor}
\def\BibTeX{{\rm B\kern-.05em{\sc i\kern-.025em b}\kern-.08em
    T\kern-.1667em\lower.7ex\hbox{E}\kern-.125emX}}

\usepackage[capitalize]{cleveref} 

\title{\LARGE \bf Long-Reach Robotic Cleaning for Lunar Solar Arrays}

\author{Stanley Wang*, Velin Kojouharov*, Long Yin Chung, Daniel Morton, and Mark Cutkosky%
\thanks{* Equal contribution.}%
\thanks{All authors are with the Dept. of Mechanical Engineering, Stanford University, Stanford, CA 94305, USA.}%
\thanks{Corresponding author email: {\tt\small swang11@stanford.edu}}}

\begin{document}

\maketitle

\section{Introduction}

Commercial lunar activity (e.g., NASA CLPS \cite{nasa_clps_2025}) is accelerating the need for reliable surface infrastructure and routine operations to keep it functioning \cite{belvin2016space}. Maintenance tasks such as inspection, cleaning, dust mitigation, and minor repair are essential to preserve performance and extend system life \cite{arney2024space}. A specific application is the cleaning of lunar solar arrays. Solar arrays are expected to provide substantial fraction of lunar surface power and operate for months to years, supplying continuous energy to landers, habitats, and surface assets, making sustained output mission-critical \cite{ramala2025feasibility}. However, over time lunar dust accumulates \cite{grun2011lunar} on these large solar arrays which can rapidly degrade panel output, demanding regular, gentle cleaning over meter-scale workspaces with stable force interaction and minimal human involvement. These requirements apply to other lunar cleaning tasks such as cleaning lander windows and viewports, other optical surfaces, radiators and thermal-control panels \cite{abel2023lunar}. These tasks motivate the use of robotic solutions that can deliver repeatable cleaning and maintenance of lunar infrastructure. 

We propose a small mobile robot equipped with a long-reach, lightweight deployable boom (Fig. 1) with an interchangeable cleaning tool (soft, abrasion-safe pad with dust capture) to perform these meter-scale tasks. The system is intended to reach solar panels mounted meters above the surface and adapt to different array configurations, such as vertical \cite{taylor2023game}. Building on our prior work demonstrating accurate, vision-guided manipulation with a deployable composite boom \cite{wang2025longreach}, we now add a compliant wrist with distal force sensing and a velocity-based admittance control policy to regulate gentle, stable contact for cleaning flat surfaces. Below, we present preliminary benchtop results demonstrating stable force regulation (~2N normal force) with a deployable boom while following a simple vertical-motion cleaning primitive.

\begin{figure}
    \centering
    \includegraphics[width=0.95\columnwidth]{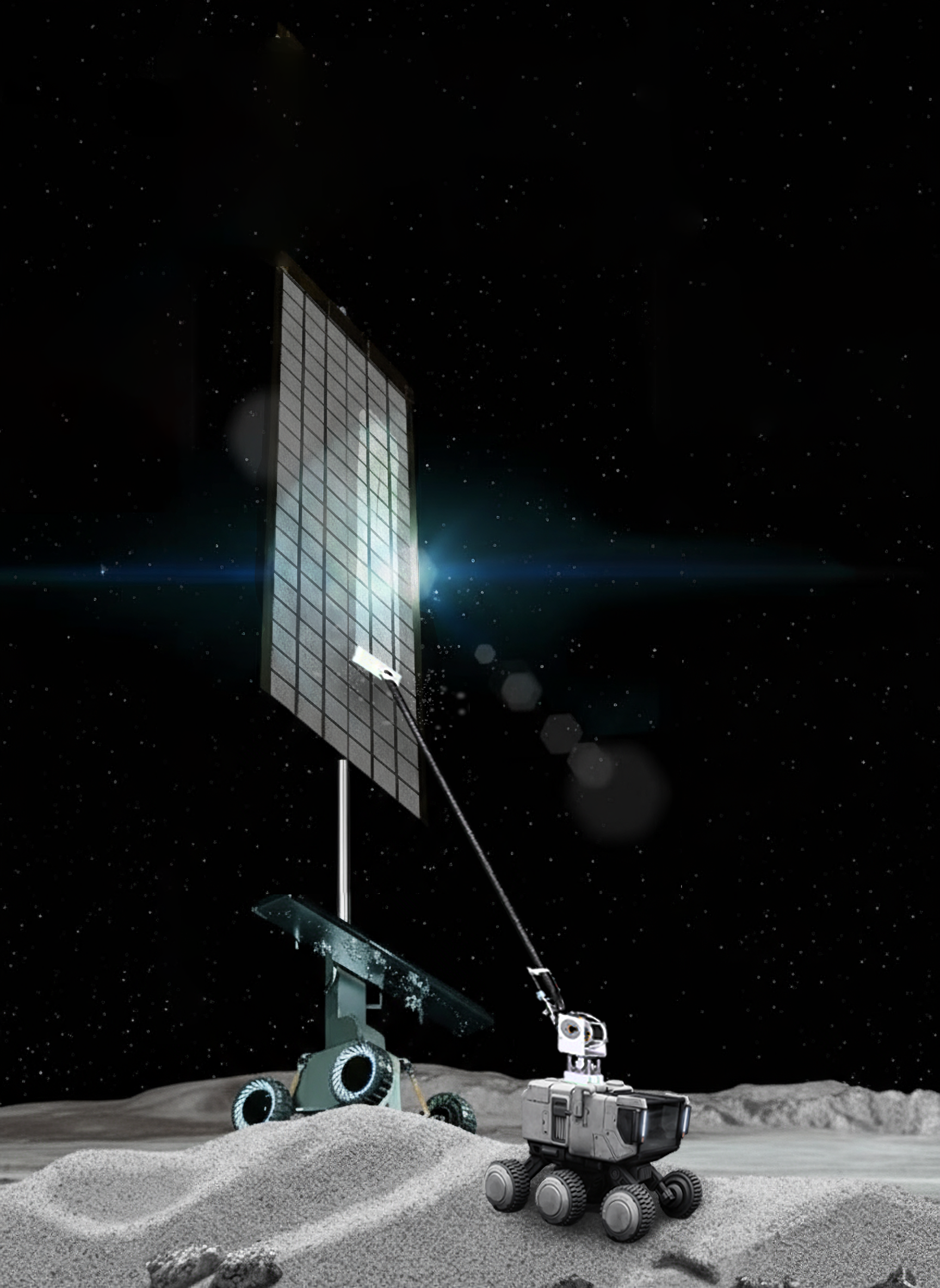}
    \caption{Concept illustration of a mobile base (wheeled rover) equipped with a long-reach (5-10m) deployable manipulator arm for cleaning large solar arrays. We acknowledge the use of AI for image generation and took inspiration from conceptual art developed by Lockheed Martin \cite{lockheed2022moonBright}.}
    \label{fig:concept}
\end{figure}

\section{Related Work}

Cleaning large, smooth surfaces, specifically solar panels is not a new problem. On Earth, solar panel maintenance relies on water/soap rinses, brush/roller systems, air jets, and manual crews \cite{patil2017review} --- approaches that are impractical on the Moon due to vacuum, limited manufacturing infrastructure, and water scarcity. Current lunar alternatives include texturing panels to reduce adhesion and active methods such as mechanical vibrations \cite{rittenhouse2025lunar} and electrostatics \cite{kawamoto2018practical}. One leading non-contact option, Electrodynamic Dust Shield (EDS), uses electrodes in the panel surface and applies phased high voltage to actively mitigate dust adhesion and accumulation on solar panels, radiators, mirrors, and other surfaces \cite{kawamoto2025electrodynamic}. While these strategies are effective for baseline dust control on prepared substrates, a mobile long-reach cleaning robot avoids the need for integrated dust-mitigation hardware, can service many arrays with one system, extends to other surfaces (radiators, viewports, instruments), and can remove heavy, event-driven residue (e.g., drilling/plume deposits) that coatings or EDS can’t handle.

However, designing and controlling such long-reach manipulators for mobile robots is challenging due to weight and packaging constraints. For the design presented in this work, we draw inspiration from deployable composite booms, traditionally used in space as single-deployment structural members for antennae, deorbit sails \cite{stohlman2013deorbitsail}, and solar arrays \cite{straubel2011deployable}. Some examples include CTM (Collapsible Tube Mast) \cite{fernandez2018bistable}, STEM (Storable Tubular Extendible Member) \cite{Thomson1993DeployableAR}, and TRAC (Triangular Rollable and Collapsible) \cite{murphey2017trac}. Roboticists have also recognized the utility of deployable structures in enabling lightweight long-reach manipulation, dating back to the Viking program rover which used a deployable arm for excavating regolith \cite{Viking}. More recent deployable robot designs include booms, masts, and everting vine robots intended for manipulation \cite{collins2016design, zippermast2015, blumenschein2020design}. While, the use of deployable booms for lunar exploration and manipulation has been proposed \cite{chen2024locomotion}, there still is limited work that shows how these deployable structures could be used in precise, force-dependent manipulation tasks, such as cleaning or inspection. This work aims to fill those gaps by building on a previous lightweight deployable-boom design \cite{wang2025longreach} and adding a force-aware control strategy that builds on admittance control methods for compliant robots \cite{seraji1994adaptive}.

\section{Control}

\begin{figure}
    \centering
    \includegraphics[width=0.95\linewidth]{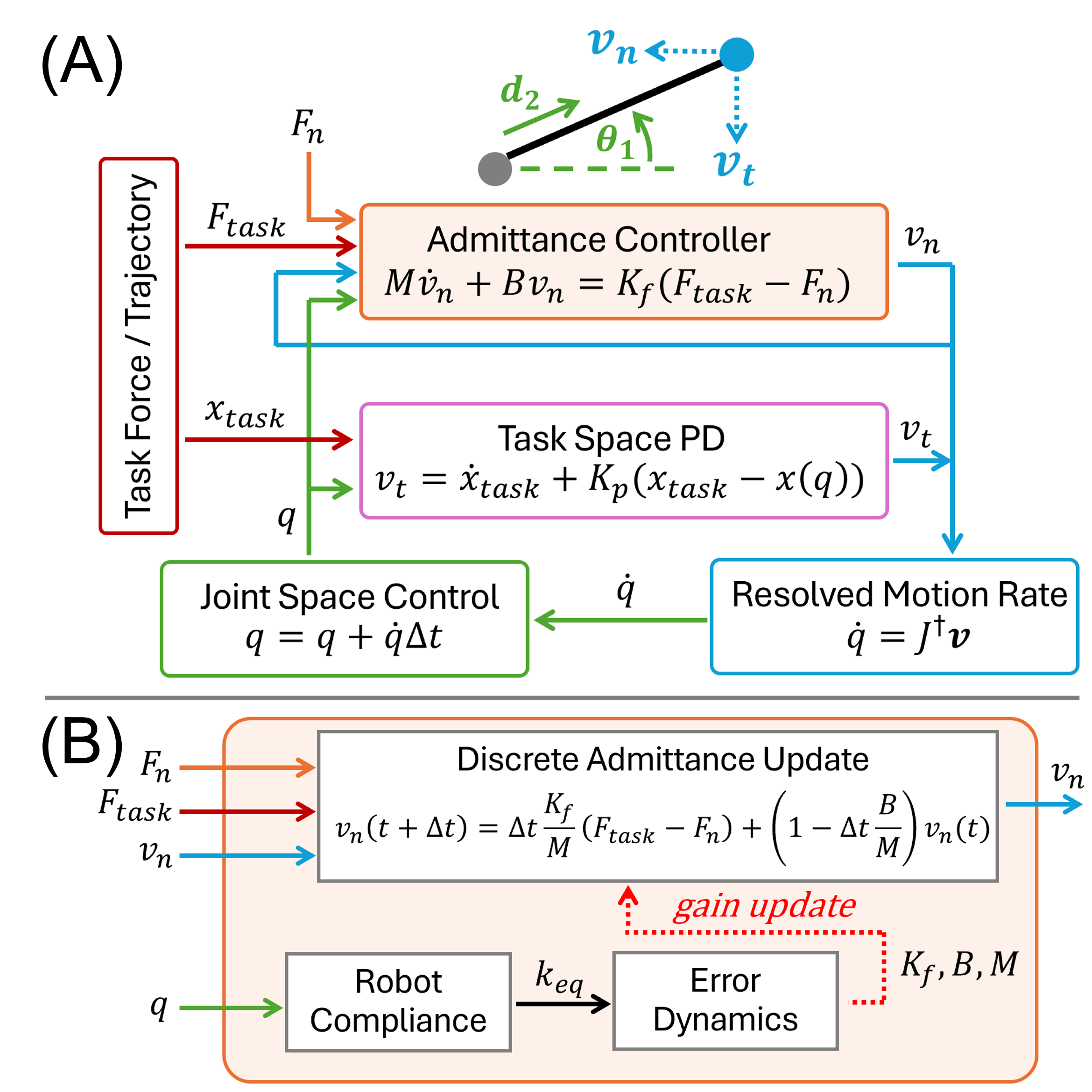}
    \caption{Control overview: measured normal force $F_n$
goes through a simple admittance control block to set approach normal velocity $v_n$. Tangential velocity $v_t$ is commanded separately for surface following. Distal force sensing defines the surface normal, so the robot keeps gentle, steady contact while it moves along the panel.}
    \label{fig:control}
\end{figure}

Fig. \ref{fig:control} illustrates an overview of the control stack. With respect to a task frame (surface), endpoint velocities in the normal $(v_n)$ and tangential $(v_t)$ directions are computed to control contact force and trajectory respectively. 
Normal contact force is modulated with velocity admittance (with gains dynamically adjusted to external task compliance).
Task velocity is then resolved to joint space with a resolved motion rate controller $(\dot{q} = J^\dagger v)$. 

\subsection{Robot Model}
To demonstrate our admittance control strategy, consider a 2 DoF (RP) long-reach manipulator (with a pitch joint $\theta_1$ controlling boom angle and an extensible boom $d_2$). This easily generalizes to a 3D case (RRP), where we simply add consideration for an additional boom angle.

The jointspace of the manipulator is thus:
\[\mathbf{q} = \begin{bmatrix}
    \theta_1 & d_2
\end{bmatrix}^T\]
Forward kinematics give the end-effector position as:
\[\mathbf{x} = \begin{bmatrix}
    d_2 \cos\theta_1 &
    d_2 \sin\theta_1
\end{bmatrix}^T\]

We model the overall bending compliance of the manipulator (beam elasticity, joint stiffness, etc) as a lumped torsional compliance on the pitch joint:
\[ \tau_1 = -k_\theta \; d\theta_1\]

\subsection{Task Space Compliance}
Consider performing tasks on a planar surface (wall), in this case with outwards normal $-\hat{x}$ for our 2D case. We are concerned with the stiffness of the robot in the task frame (local surface normal) in order to properly control contact force.

From forward kinematics, we establish the differentials:
\[dx = -d_2 \sin\theta_1 d\theta_1\]
Furthermore, for a normal force $F_x$ at the endpoint, the resultant torque around the base is given as:
\[\tau_1 = -F_x \sin\theta_1 d_2\]
Using this, we can transform the pitch joint stiffness to an equivalent stiffness in the task frame:
\[F_x = -\frac{k_\theta}{d_2^2 \sin^2\theta_1} dx = -k_x dx\]
We additionally consider $k_{ee}$, the normal stiffness at the contact itself (i.e. from the wall or end-effector). Combined with the manipulator stiffness, the equivalent task-space stiffness becomes:
\[k_{n} = \left(\frac{1}{k_x} +\frac{1}{k_{ee}}\right)^{-1}\]

\subsection{Task Force Control}
We utilize a velocity admittance controller for modulating task force. Given a desired normal contact force $F_d$ and measured force $F_n$, we control the normal velocity $v_n$ (into or out of the surface) subject to virtual mass and damping:
\[M \dot{v_n} + B v_n = K_f (F_d - F_n)\]
The measured force is a direct result of the task-space compliance $F_n = k_{eq} x$. We can thus write the dynamics of the error $(e = F_d - k_{eq}x)$ as:
\[M \ddot{e} + B \dot{e} + K_f k_{eq} e = 0\]
This yields a second-order system where error behavior can be tuned. Given a desired natural frequency and damping $(\omega_n, \eta)$, we can dynamically adjust the admittance gains based on the current task-space compliance:
\begin{align*}
    K_f &= \frac{\omega_n^2}{k_{eq}}M \\
    B &= 2\eta\omega_n M
\end{align*}

 Applying an Euler approximation to the admittance control law gives us the discrete update loop (500 Hz+):
 \[v_n(t+\Delta t) = \Delta t \frac{K_f}{M}(F_{task}-F_n) + \left(1 - \Delta t \frac{B}{M}\right) v_n(t)\]

\subsection{Task Trajectory Control}
Tangential velocity $v_t$ in the task frame (movements parallel to the local surface) are controlled for trajectory following. 
In cleaning tasks, trajectories may be fairly primitive (i.e. sweeping up and down), but enabling more complex trajectory control motivates further dexterous surface manipulation applications (i.e. painting). 

Given a task trajectory $x_{task}(t)$, we implement a PD controller on tangential velocity:
\[v_t = \dot{x}_{task} + K_p(x_{task} - x(q))\]
This control loop provides slower updates (\(\approx\) 10 Hz), since vision-based perception strategies (visual endpoint servoing) are utilized to estimate the position $x(q)$.

\section {Experiments}

\begin{figure}
    \centering
    \includegraphics[width=0.95\linewidth]{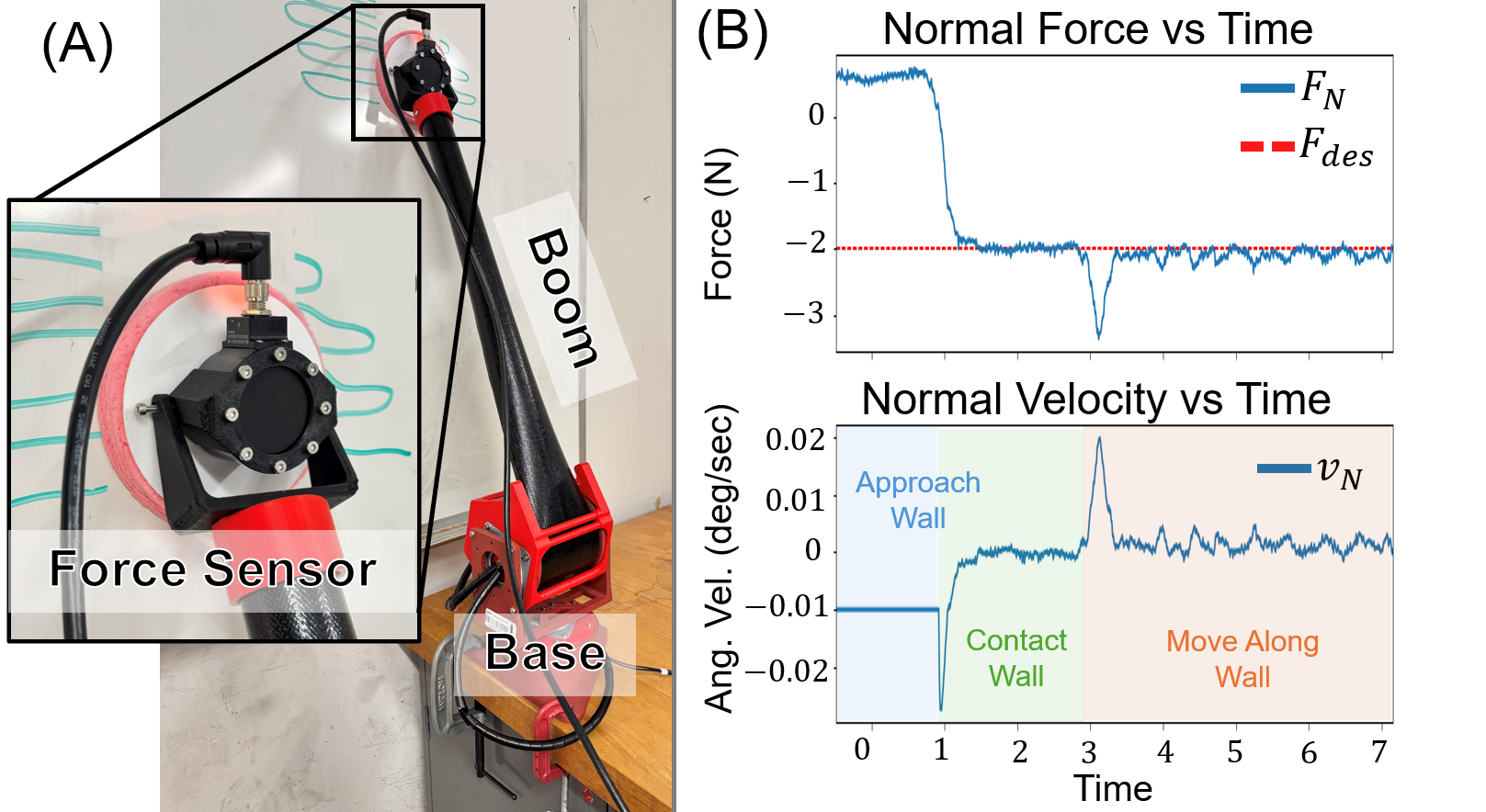}
    \caption{Experimental setup and representative trial. (A) Long-reach deployable boom with compliant self-aligning wrist, inline 6-axis force sensor, and soft,  wiper pad contacting a vertical surface. (B) Force (top) and commanded normal velocity (bottom) during approach (blue), force stabilization (green), and trajectory following (orange) with a normal-axis admittance controller tracking with desired force $F_{des}=-2N$}.
    \label{fig:experiment}
\end{figure}

To evaluate the proposed control strategy, we performed a simple, low-force test on a planar surface (a whiteboard used as a rigid, smooth proxy for a panel). We used the deployable-boom arm from prior work \cite{wang2025longreach} and added a compliant wrist with an inline 6-axis force sensor and a soft wiper pad \cref{fig:experiment}A. The compliant wrist provides passive self-alignment to the surface normal so that the measured axial load closely approximates normal force with the wall without requiring precise knowledge of wall angle.

First, the robot starts completely off of the wall and is commanded via a slow, constant velocity trajectory in the direction of the wall. This is seen in the blue region of \cref{fig:experiment}B. Then, using the force sensor, we detect initial wall contact and immediately turn on the velocity-based admittance controller below with desired force, $F_{des} = -2N$ (This force is negative because the sensor is in compression).

 \[v_n(t + \Delta t) = \Delta t \frac{K_f}{M}(F_{des}-F_n) + \left(1 - \Delta t \frac{B}{M}\right) v_n(t)\]

 This is the section seen in green in \cref{fig:experiment}B. Once the force is constant at the desired force threshold, the robot continues using the admittance controller but also moving along the surface with a constant tangent velocity up. This region is shown in orange. In general, the robot extended from a length of 0.3m to 1.0m and maintained an $\mathrm{RMS}\,\lvert F_N - F_{\mathrm{des}}\rvert \approx 0.2N$ after initial contact.

 While these results are preliminary, they do highlight the effectiveness of our controller. The robot was able to maintain the normal forces with the surface ranging from 1-10N. The main deviation from the desired force often happens right as the tangential motion begins. We believe that this is because of the stiction of the compliant pad against the whiteboard. Future work will look at ways to mitigate this effect. 

\section{Future Work and Discussion}

In the future, we plan to generalize our controller to 3D contact on tilted surfaces and augment with a mobile base.  Our bench-top results indicate that a deployable, length-varying arm can sustain few-newton contact with a simple normal-axis admittance loop while decoupling tangential motion. At longer reach (multiple meters), low torsional stiffness and elastic bending create low-frequency force vibrations and in the future, we plan to build simulations to better model these effects, especially in lower lunar gravity. Also, we have only demonstrated simple linear trajectories up and down a surface. While this is a relatively standard way to clean a surface, we hope to extend to more sophisticated trajectories along the surface we are cleaning. This will likely require endpoint visual servoing. Additionally, while this work uses a fixed-base arm, we envision these long reach manipulators deployed from legged or wheeled mobile platforms on the lunar surface. We will have to consider base stability  and contact-aware whole body coordination \cite{khatib2008unified} and stance planning \cite{morton2024reachbot}. Together, these steps will extend the current method from planar, 2D wall tests to 3D surface following on a more complete robot embodiment while preserving gentle, regulated contact.

\section*{Acknowledgment}
\emph{Special Thanks:} The Stanford Robotics Center (SRC), including Matt Van Cleave, Eiko Rutherford, Zen Yaskawa, Steve Cousins, and Oussama Khatib.

\emph{Funding:} Stanley Wang and Venny Kojouharov were supported by the NSF Graduate Research Fellowship (with additional support for Venny from Knight-Hennessy Scholars), and Daniel Morton was supported by a NASA Space Technology Graduate Research Opportunity.

\emph{AI Use:} utilized DALL-E (OpenAI) for generation of the background image and rover asset. These were then composited with CAD renders of our manipulator arm.

\bibliographystyle{IEEEtran}
\bibliography{citations}

@misc{nasa_clps_2025,
  author       = {{National Aeronautics and Space Administration}},
  title        = {Commercial Lunar Payload Services},
  howpublished = {\url{https://www.nasa.gov/commercial-lunar-payload-services/}},
  note         = {Accessed: 1 August 2025},
  year         = {2025}
}

@article{arney2024space,
  title={In-space Servicing, Assembly, and Manufacturing (ISAM) State of Play-2024 Edition},
  author={Arney, Dale and Mulvaney, John and Williams, Christina and Hernandez, Wilbert Andres Ruperto and Friz, Jessica and Stockdale, Christopher and Nelson, John and Vargas, Rafael Rivera},
  year={2024}
}

@inproceedings{belvin2016space,
  title={In-space structural assembly: Applications and technology},
  author={Belvin, Wendel K and Doggett, William R and Watson, Judith J and Dorsey, John T and Warren, Jay E and Jones, Thomas C and Komendera, Erik E and Mann, Troy and Bowman, Lynn M},
  booktitle={3rd AIAA Spacecraft Structures Conference},
  pages={2163},
  year={2016}
}

@inproceedings{collins2016design,
  title={Design of a spherical robot arm with the spiral zipper prismatic joint},
  author={Collins, Foster and Yim, Mark},
  booktitle={2016 IEEE international conference on robotics and automation (ICRA)},
  pages={2137--2143},
  year={2016},
  organization={IEEE}
}

@misc{zippermast2015,
  author       = {{I. Geo Systems}},
  title        = {ZM-20 Zippermast},
  howpublished = {\url{http://zippermast.com/sam-15/}},
  note         = {Accessed: 2015-09-01},
  year         = {2015}
}

@article{blumenschein2020design,
  title={Design, modeling, control, and application of everting vine robots},
  author={Blumenschein, Laura H and Coad, Margaret M and Haggerty, David A and Okamura, Allison M and Hawkes, Elliot W},
  journal={Frontiers in Robotics and AI},
  volume={7},
  pages={548266},
  year={2020},
  publisher={Frontiers Media SA}
}

@article{chen2024locomotion,
  title={Locomotion as manipulation with reachbot},
  author={Chen, Tony G and Newdick, Stephanie and Di, Julia and Bosio, Carlo and Ongole, Nitin and Lap{\^o}tre, Mathieu and Pavone, Marco and Cutkosky, Mark R},
  journal={Science Robotics},
  volume={9},
  number={89},
  pages={eadi9762},
  year={2024},
  publisher={American Association for the Advancement of Science}
}

@inproceedings{straubel2011deployable,
  title={Deployable composite booms for various gossamer space structures},
  author={Straubel, Marco and Block, Joachim and Sinapius, Michael and H{\"u}hne, Chrisitan},
  booktitle={52nd AIAA/ASME/ASCE/AHS/ASC Structures, Structural Dynamics and Materials Conference 19th AIAA/ASME/AHS Adaptive Structures Conference 13t},
  pages={2023},
  year={2011}
}

@inproceedings{stohlman2013deorbitsail,
  title={Deorbitsail: a deployable sail for de-orbiting},
  author={Stohlman, Olive R and Lappas, Vaios},
  booktitle={54th AIAA/ASME/ASCE/AHS/ASC Structures, Structural Dynamics, and Materials Conference},
  pages={1806},
  year={2013}
}

@inproceedings{morton2024reachbot,
        author={Morton, Daniel and Cutkosky, Mark and Pavone, Marco},
        booktitle={2024 IEEE/RSJ International Conference on Intelligent Robots and Systems (IROS)}, 
        title={Task-Driven Manipulation with Reconfigurable Parallel Robots}, 
        year={2024},
        pages={9924-9930},
        doi={10.1109/IROS58592.2024.10801313},
      }

@inproceedings{Thomson1993DeployableAR,
  title={Deployable and retractable telescoping tubular structure development},
  author={Mark Thomson},
  year={1993},
  url={https://api.semanticscholar.org/CorpusID:109203948}
}

@inproceedings{murphey2017trac,
  title={TRAC boom structural mechanics},
  author={Murphey, Thomas W and Turse, Dana and Adams, Larry},
  booktitle={4th AIAA spacecraft structures conference},
  pages={0171},
  year={2017}
}

@inproceedings{fernandez2018bistable,
  title={Bistable collapsible tubular mast booms},
  author={Fernandez, Juan M and Lee, Andrew J},
  booktitle={International Conference on Advanced Lightweight Structures and Reflector Antennas},
  number={NF1676L-30217},
  year={2018}
}

@article{Viking,
  title={The viking project},
  author={Soffen, Gerald A},
  journal={Journal of Geophysical Research},
  volume={82},
  number={28},
  pages={3959--3970},
  year={1977},
  publisher={Wiley Online Library}
}

@inproceedings{khatib2008unified,
  title={A unified framework for whole-body humanoid robot control with multiple constraints and contacts},
  author={Khatib, Oussama and Sentis, Luis and Park, Jae-Heung},
  booktitle={European Robotics Symposium 2008},
  pages={303--312},
  year={2008},
  organization={Springer}
}

@article{kawamoto2018practical,
  title={Practical performance of an electrostatic cleaning system for removal of lunar dust from optical elements utilizing electrostatic traveling wave},
  author={Kawamoto, Hiroyuki and Hashime, Shusuke},
  journal={Journal of Electrostatics},
  volume={94},
  pages={38--43},
  year={2018},
  publisher={Elsevier}
}

@article{rittenhouse2025lunar,
  title={Lunar dust mitigation for solar cells via ultrasonic vibrations},
  author={Rittenhouse, Jeremiah J and Boeringa, Zachary L and Han, Daoru and Stutts, Daniel S},
  journal={Acta Astronautica},
  volume={228},
  pages={474--485},
  year={2025},
  publisher={Elsevier}
}

@inproceedings{patil2017review,
  title={A review on cleaning mechanism of solar photovoltaic panel},
  author={Patil, PA and Bagi, JS and Wagh, MM},
  booktitle={2017 International Conference on Energy, Communication, Data Analytics and Soft Computing (ICECDS)},
  pages={250--256},
  year={2017},
  organization={IEEE}
}

@article{abel2023lunar,
  title={Lunar Dust Mitigation: A Guide and Reference},
  author={Abel, PB and others},
  journal={National Aeronautics and Space Administration, NASA/TP-20220018746},
  volume={1},
  year={2023}
}

@inproceedings{taylor2023game,
  title={Game Changing Development Program-Vertical Solar Array Technology (VSAT) Project},
  author={Taylor, Chuck and Pappa, Richard},
  booktitle={Game Changing Development Annual Program Review},
  year={2023}
}

@inproceedings{ramala2025feasibility,
  title={A Feasibility Study on Space-Based Solar Power for Lunar Economy},
  author={Ramala, Sumanth Kumar Reddy and Garzaniti, Nicola},
  booktitle={AIAA AVIATION FORUM AND ASCEND 2025},
  pages={4027},
  year={2025}
}

@article{grun2011lunar,
  title={The lunar dust environment},
  author={Gr{\"u}n, Eberhard and Horanyi, Mihaly and Sternovsky, Zoltan},
  journal={Planetary and Space Science},
  volume={59},
  number={14},
  pages={1672--1680},
  year={2011},
  publisher={Elsevier}
}

@article{kawamoto2025electrodynamic,
  title={Electrodynamic dust removal technologies for solar panels: A comprehensive review},
  author={Kawamoto, Hiroyuki},
  journal={Journal of Electrostatics},
  pages={104045},
  year={2025},
  publisher={Elsevier}
}

@inproceedings{seraji1994adaptive,
  title={Adaptive admittance control: An approach to explicit force control in compliant motion},
  author={Seraji, Homayoun},
  booktitle={Proceedings of the 1994 IEEE International Conference on Robotics and Automation},
  pages={2705--2712},
  year={1994},
  organization={IEEE}
}

@inproceedings{wang2025longreach,
  title        = {Long-Reach Robotic Manipulation for Assembly and Outfitting of Lunar Structures},
  author       = {Wang, Stanley and Kojouharov, Venny and Chung, Long Yin and Morton, Daniel and Cutkosky, Mark},
  booktitle    = {2025 International Symposium on Space Robotics (iSpaRO)},
  pages        = {xx--xx},
  year         = {2025},
  organization = {IEEE}
}

@online{lockheed2022moonBright,
  author    = {{Lockheed Martin}},
  title     = {Humanity's Future on the Moon is Looking Bright},
  year      = {2022},
  month     = oct,
  day       = {24},
  url       = {https://www.lockheedmartin.com/en-us/news/features/2022/humanity-s-future-on-the-moon-is-looking-bright.html},
  note      = {Feature article},
  urldate   = {2025-10-31}
}


\end{document}